\documentclass[10pt,twocolumn,letterpaper]{article}

\usepackage{cvpr}              
\definecolor{cvprblue}{rgb}{0.21,0.49,0.74}
\usepackage[pagebackref,breaklinks,colorlinks,allcolors=cvprblue]{hyperref}
\usepackage{xcolor}
\usepackage{xspace}
\newcommand{\method}{LoGo\xspace} 
\usepackage{booktabs}
\usepackage{graphicx}
\usepackage{enumitem}
\newcommand{\meanstd}[2]{\ensuremath{#1{\pm #2}}}
\newcommand{\bestmeanstd}[2]{\ensuremath{\mathbf{#1{\pm #2}}}}
\newcommand{\secondmeanstd}[2]{\ensuremath{\underline{#1{\pm #2}}}}

\def\confName{CVPR}
\def\confYear{2027}

\title{
LoGo: Look-to-Go via Task-Aware Active Perception \\ for End-to-End Legged Mobility
}

\author{First Author\\
Institution1\\
Institution1 address\\
{\tt\small firstauthor@i1.org}
\and
Second Author\\
Institution2\\
First line of institution2 address\\
{\tt\small secondauthor@i2.org}
}
\author{
Dapeng Jia$^{1,*}$ \quad
LI Peizhuo$^{1,*}$ \quad
Hongyi Li$^{2}$ \quad
Yujie Chen$^{3}$ \quad
Zhiheng Zou$^{3}$ \quad
Weilin Xia$^{3}$ \quad
Bida Ma$^{3}$ \\
Zhiyou Yu$^{3}$ \quad
Zhilin Xiong$^{3}$ \quad
Fangzhou Xu$^{1}$ \quad
Qiwei Li$^{4}$ \quad
Yueqi Zhang$^{3}$ \quad
Zhihan Zhu$^{3}$ \quad
Ling Xiao$^{3}$ \\ 
Yifeng Zhang$^{1,\dagger}$ \quad
Mingfeng Fan$^{1}$ \quad 
Annan Tang$^{3}$ \quad
Armin Lederer$^{1}$ \quad
Guillaume Sartoretti$^{1}$ \\[0.8em]
$^{1}$National University of Singapore \quad 
$^{2}$Zhejiang University \\ 
$^{3}$EngineAI Robotics Technology Co., Ltd. \quad 
$^{4}$Peking University \\[0.5em]
{\small
$^{*}$Equal contribution.
\quad
$^{\dagger}$Corresponding author.
}
}

\begin{document}

\twocolumn[{
    \renewcommand\twocolumn[1][]{#1}
    \maketitle
        \vspace{-0.3cm}
    \noindent
    \begin{minipage}{\textwidth}
        \centering
        \includegraphics[width=\linewidth]{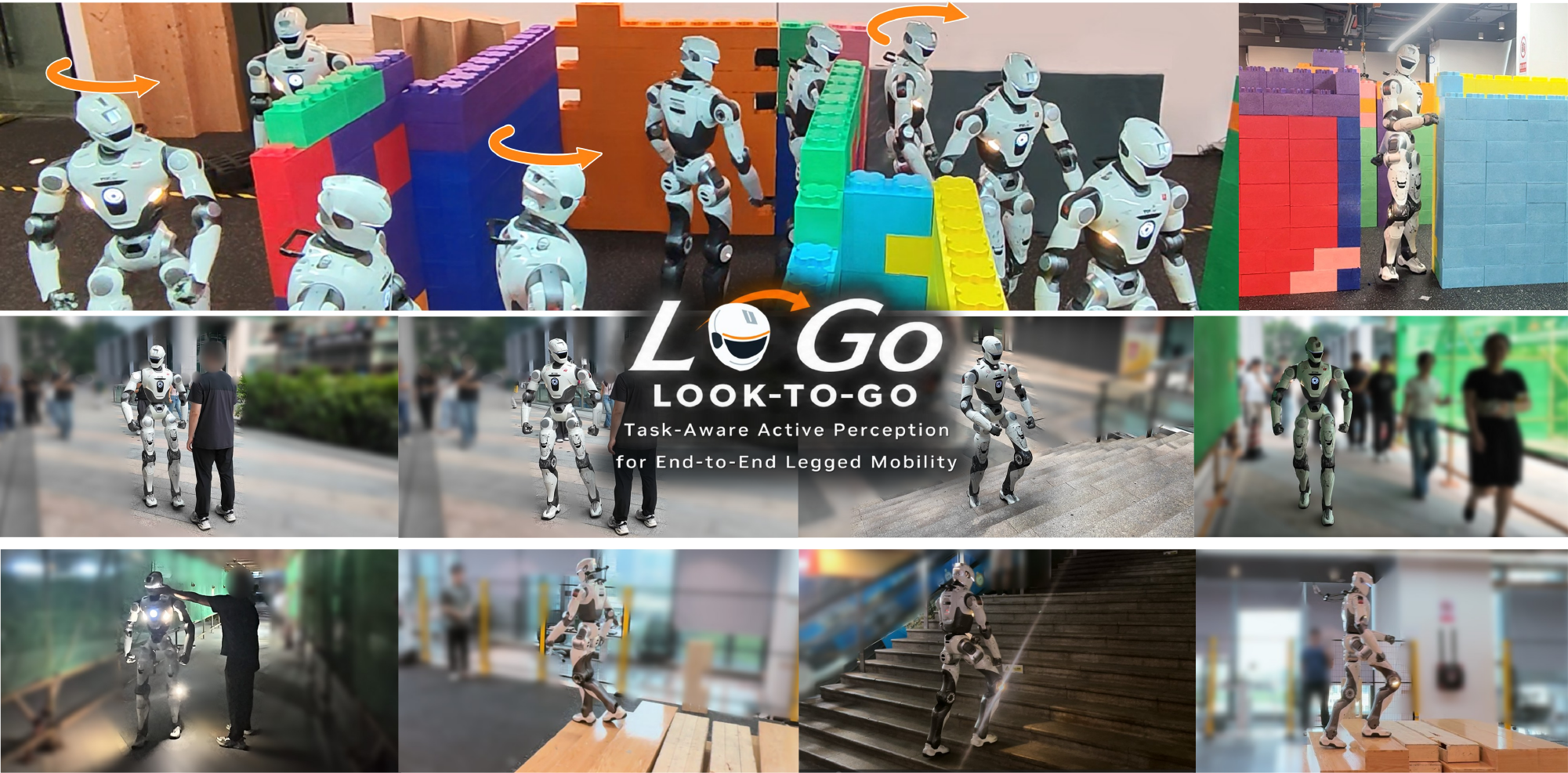}
        \captionof{figure}{
        LoGo learns to coordinate task-aware active visual perception, goal-directed navigation, and whole-body locomotion within a single end-to-end policy, enabling adaptive humanoid mobility across diverse and complex real-world environments.
}   
        \label{fig:intro}
    \end{minipage}
    \par\vspace{1em}
}]

\begin{abstract}
Humans actively seek out task-relevant visual information, directing their gaze toward regions that can inform subsequent actions.
In contrast, robotic locomotion often relies on fixed or predefined perceptual strategies, limiting its ability to anticipate terrain changes and explore alternative paths during movement.
To address this limitation, we introduce LoGo (Look-to-Go), a bio-inspired, end-to-end reinforcement learning framework that jointly learns task-aware active visual perception, goal-directed navigation, and whole-body locomotion within a single policy.
By reasoning over its complementary spatial memory, body state, and immediate surrounding, our framework allows the robot to learn anticipatory visual behaviors that emerge from task-driven learning.
This allows the robot to direct its gaze to proactively inspect potential routes, navigate around obstacles, and even adapt its motion to fit through narrow passages, without explicit viewpoint supervision or observation-specific rewards.
Our results suggest such task-driven coupled perception and locomotion strategy may provide a foundation for more autonomous and adaptive embodied intelligence.
\end{abstract}

\vspace{-0.6cm}
\section{Introduction}
\vspace{-0.3cm}
\label{sec:introduction}
Legged robots are increasingly expected to operate in human-centered urban environments for tasks such as delivery, inspection, and daily assistance~\cite{ginting2024seek,cai2024navigating}. 
These applications require legged mobility: the ability to integrate goal-directed navigation with whole-body locomotion to traverse complex terrain and overcome obstacles~\cite{lee2024learning}.
Yet the visual information needed to guide upcoming actions shifts with task progress, terrain structure, and changes in the environment. 
For example, the robot may need to observe the steps ahead during stair climbing or a side passage before a turn.
Limited fields of view and frequent occlusions prevent any single viewing direction from capturing all relevant information. 
\emph{Where to look} therefore becomes a third decision, tightly coupled with \emph{where to go} and \emph{how to move}.

Existing approaches address the first two decisions but leave the third largely unaddressed.
Hierarchical methods separate navigation from locomotion and connect them through intermediate commands~\cite{kareer2023vinl,jeong2026hipan,yang2026spatially}, while end-to-end approaches unify them by directly mapping visual and proprioceptive inputs to joint actions~\cite{agarwal2023legged,cheng2024extreme,li2025kivi}.
Both rely on body-fixed sensors, so the viewing direction is tied to body orientation, and the robot can only act on what happens to be in view.
Active perception methods do control the viewpoint, but they typically depend on predefined sensing objectives, information-gain rewards, or viewpoint supervision~\cite{malczyk2026reinforcement,xiong2025vision}, while task-relevant attention methods only select information from already available observations without acquiring new views~\cite{li2026taga}.
As a result, \emph{where to look} is either fixed or manually specified, rather than learned together with \emph{where to go} and \emph{how to move}.

In biological locomotion, by contrast, these three decisions are tightly coordinated rather than made separately. Humans and other animals shift their gaze toward the upcoming path before the body turns, so that \emph{where to look} both anticipates and informs \emph{where to go} and \emph{how to move}~\cite{hollands2002look,ros2017pigeons}.
This motivates us to ask whether the same coordination can emerge in a legged robot whose viewing direction is controlled independently of its body, e.g., via an actuated head, without explicitly prescribing where to look.
To this end, we propose \method (Look-to-Go), an end-to-end reinforcement learning (RL) framework that jointly learns task-aware active perception, goal-directed navigation, and whole-body locomotion within a single policy.

The architecture of \method follows a functional division found in biological visually guided locomotion, where movement relies on an internal estimate of body state, a memory of the environment, and a body-centered representation of nearby space~\cite{mcvea2006long}.
Accordingly, \method builds three complementary representations.
\emph{Body State} maintains an internal estimate of the robot's motion and contact states from proprioception, analogous to the internal models the nervous system uses to estimate and predict the state of the body~\cite{wolpert1995internal}.
\emph{Spatial Memory} retains previously observed terrain that has left the field of view, analogous to the memory of obstacles that walking animals use to guide subsequent steps~\cite{mcvea2006long,joshi2023dynamic}.
\emph{Immediate Surrounding} represents current visual observations relative to the robot's body, analogous to the body-centered encoding of nearby space in the posterior parietal cortex~\cite{marigold2017posterior}.
Conditioned on these representations, a single policy outputs joint-space actions and is trained end-to-end with RL, so that active perception, navigation, and locomotion are coordinated without viewpoint supervision or dedicated observation rewards. 

Our main contributions are summarized as follows:

\begin{enumerate}
    \item We introduce LoGo, an end-to-end RL framework that jointly learns active perception, navigation, and whole-body locomotion without explicit viewpoint supervision or observation-specific rewards.
    \item We design a bio-inspired perception architecture that builds three complementary representations, namely Body State, Spatial Memory, and Immediate Surrounding, to support coordinated perception and action.
   
    \item We demonstrate robust mobility across diverse terrains and challenging environments. In simulation, \method is validated across embodiments on both humanoid and quadruped robots; in the real world, it is deployed on a full-size humanoid robot in narrow passages, pedestrian-populated areas, and changing outdoor conditions.
    
    \item We show that task-aware active perception emerges from the mobility objective alone: without any viewing supervision, our learned policy directs its view according to the goal and looks toward the path before the body turns, consistent with gaze behavior in human locomotion.
   
\end{enumerate}

\vspace{-0.1cm}
\section{Related Work}
\vspace{-0.1cm}
\label{sec:related_work}
\noindent\textbf{Vision-Based Legged Mobility.}
Hierarchical approaches often connect navigation and locomotion through velocity, posture, or skill commands~\cite{kareer2023vinl,jeong2026hipan,hoeller2024anymal,cheng2024navila,wang2026guide}.
In contrast, end-to-end policies map egocentric vision and proprioception to joint actions~\cite{agarwal2023legged,cheng2024extreme}.
Visual parkour methods enable traversal of challenging terrain~\cite{zhuang2023robot,chane2024soloparkour,rudin2025parkour,zhuang2024humanoid,zhu2026hiking}, while goal-conditioned controllers integrate terrain traversal with obstacle avoidance~\cite{he2024agile,zhang2026ame}.
However, body-fixed cameras constrain viewing direction to body orientation, requiring the robot to reorient its body to look elsewhere.

\noindent\textbf{Attention and Memory in Legged Locomotion.}
Proprioceptive information guides attention over terrain maps~\cite{he2025attention}, while TAGA selects local height-scan regions for velocity tracking~\cite{li2026taga}.
KiVi combines visual and body-state representations through temporal attention~\cite{li2025kivi}, while neural volumetric memory accumulates spatially aligned visual features~\cite{yang2023neural}.
These methods focus on processing available observations, whereas actively redirecting the sensor determines what the robot can observe next.

\noindent\textbf{Task-Driven Active Vision.}
\begin{figure*}[t]
    \includegraphics[width=\linewidth]{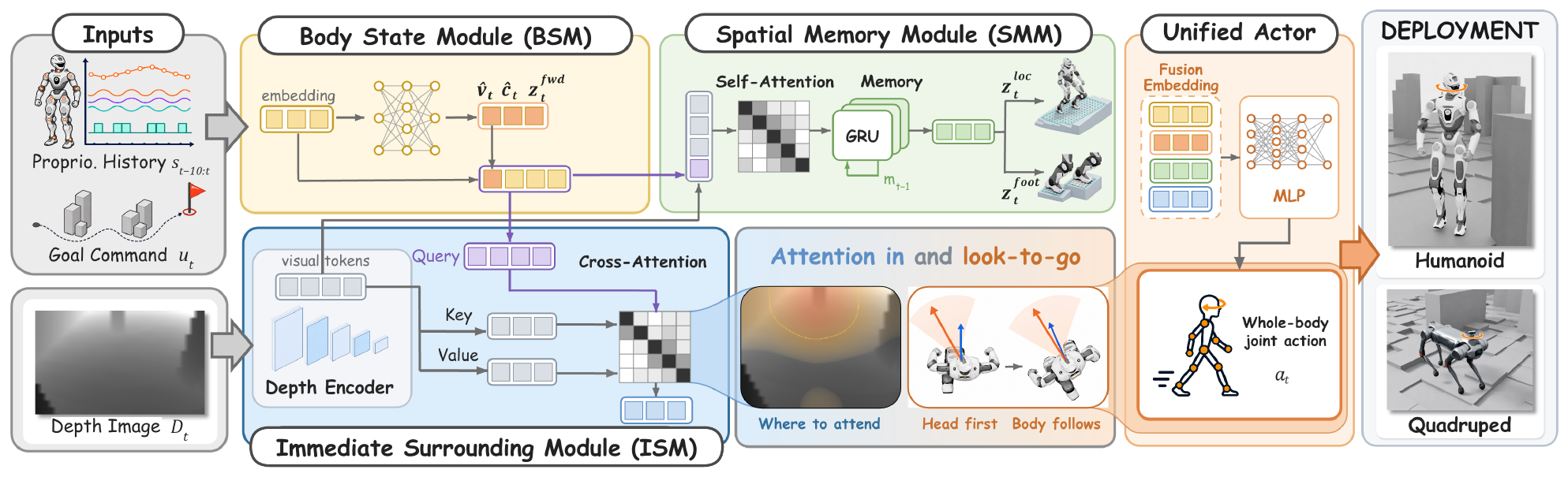}
    \caption{\textbf{Overview of the LoGo framework.} Body State, Spatial Memory, and Immediate Surrounding modules fuse body-state estimates, visual history, and body-conditioned features in a unified policy for task-aware active perception and goal-directed legged mobility.}
    \label{fig:framework}
    \vspace{-0.5cm}
\end{figure*}
Recent studies have explored active vision across manipulation and navigation tasks.
In manipulation, gaze behaviors have been learned through imitation~\cite{kerr2025eye,xiong2025vision,yu2026egomi}, hierarchical fixation~\cite{hari2026eyerobot}, and world-action models~\cite{wu2026activewam}, while TAVIS investigates the temporal coordination between gaze and action~\cite{spigler2026tavis}.
Beyond manipulation, active sensing has been studied using intrinsic sensorimotor signals~\cite{shang2023active} and information-based rewards for aerial navigation~\cite{malczyk2026reinforcement}.
Other approaches investigate viewpoint selection for panoramic search~\cite{yu2026thinking} and social navigation~\cite{vice2026active}.
More closely related to our work, LEAP jointly predicts camera orientation and body velocity commands using navigation rewards, with a separate locomotion policy tracking the predicted commands~\cite{gokbakan2026leap}.
In contrast, LoGo directly learns viewpoint selection and whole-body joint control within a single policy, driven solely by task rewards.

\vspace{-0.2cm}
\section{Learning to Look-to-Go}
\label{sec:method}
We formulate task-aware legged mobility as a unified perception-action learning problem (Sec.~\ref{sec:method_problem}), in which an embodied agent jointly decides where to look, where to go, and how to move in order to accomplish its task. 
As illustrated in Fig.~\ref{fig:framework}, \method maps egocentric visual observations, proprioceptive histories, and a goal command to the actions of both the body and an actuated viewing direction through a single end-to-end policy. 
The policy builds on three representation modules, namely a Body State Module (BSM), a Spatial Memory Module (SMM), and an Immediate Surrounding Module (ISM) (Sec.~\ref{sec:method_arch}), and is trained in three progressive stages with multi-critic RL and auxiliary objectives with privileged supervision (Sec.~\ref{sec:method_training}).
\vspace{-0.2cm}
\subsection{Problem Formulation}
\label{sec:method_problem}
\paragraph{Task.}
In each episode, the robot must reach a goal position on cluttered terrain, typically one whose straight-line path is blocked by obstacles (Sec.~\ref{sec:method_env}). Since the robot perceives its surroundings only through egocentric vision, we model the problem as a partially observable Markov decision process, in which the policy must infer the unobserved terrain, its own velocity, and its contact state from onboard observations and their history.
\vspace{-0.5cm}
\paragraph{Observations.}
At each time step $t$, the policy receives the proprioceptive history $\mathbf{s}_{t-10:t}$ together with an egocentric depth image $\mathbf{D}_t$ acquired by the head-mounted sensor.
The observation vector is defined as $\mathbf{s}_t = [\boldsymbol{\omega}_t, \mathbf{g}_t, \mathbf{u}_t, \mathbf{q}_t, \dot{\mathbf{q}}_t, \mathbf{a}_{t-1}]$.
Here, $\boldsymbol{\omega}_t \in \mathbb{R}^{3}$ is the base angular velocity, $\mathbf{g}_t \in \mathbb{R}^{3}$ represents the projected gravity vector, and $\mathbf{u}_t \in \mathbb{R}^{2}$ specifies the goal command.
The vectors $\mathbf{q}_t, \dot{\mathbf{q}}_t \in \mathbb{R}^{23}$ contain the joint positions and velocities, respectively, while $\mathbf{a}_{t-1} \in \mathbb{R}^{23}$ denotes the action applied at the previous time step.
The goal command is computed from the planar goal offset $\mathbf{d}_t \in \mathbb{R}^{2}$, expressed in the robot frame, as
$
\mathbf{u}_t = \mathbf{d}_t / \max\left(1, \|\mathbf{d}_t\|_2\right).
$
This yields a unit direction vector when the goal is more than 1\,m away and retains the planar offset during the final approach.
The base linear velocity $\mathbf{v}_t$ is not directly observed and is estimated from the proprioceptive history by the Body State Module.
\vspace{-0.5cm}
\paragraph{Actions and active viewing.}
The action $\mathbf{a}_t \in \mathbb{R}^{23}$ specifies target angles for 23 joints of the humanoid, namely the 22 leg and arm joints and the head-yaw joint, which are tracked by PD controllers. 
As the sensor is rigidly mounted on the head, turning the head changes the viewing direction.
\vspace{-0.5cm}
\subsection{Policy Architecture}
\label{sec:method_arch}
As shown in Fig.~\ref{fig:framework}, our policy comprises three complementary representation modules. These representations are integrated by a unified actor to jointly determine body movements and viewing direction. 
\vspace{-0.2cm}
\paragraph{BSM.}
Inspired by the internal models that the nervous system maintains to estimate and predict body states~\cite{wolpert1995internal}, the BSM infers the robot's state from proprioceptive history. Given $\mathbf{s}_{t-10:t}$, it produces two explicit estimates and one latent predictive representation:
\begin{equation}
  \left[
    \hat{\mathbf{v}}_t,\,
    \hat{\mathbf{c}}_t,\,
    \mathbf{z}^{\text{fwd}}_t
  \right]
  = f_{\text{BSM}}\!\left(\mathbf{s}_{t-10:t}\right),
  \label{eq:body_encoding}
\end{equation}
where $f_{\text{BSM}}$ consists of an MLP-based history encoder and output heads, $\hat{\mathbf{v}}_t$ is the estimated base linear velocity, $\hat{\mathbf{c}}_t$ contains the foot-contact probabilities, and $\mathbf{z}^{\text{fwd}}_t$ serves as a predictive body-state representation.
The BSM is trained using privileged simulation signals with the auxiliary loss:
{\small
\begin{equation}
  \mathcal{L}_{\text{body}} =
  \|\hat{\mathbf{v}}_t-\mathbf{v}_t\|_2^2
  + \mathrm{BCE}(\hat{\mathbf{c}}_t,\mathbf{c}_t)
  + \|g_{\text{fwd}}(\mathbf{z}^{\text{fwd}}_t)
  - \mathbf{s}_{t+1}\|_2^2,
  \label{eq:body_State}
\end{equation}
}
\noindent where $\mathbf{v}_t$ and $\mathbf{c}_t$ denote the ground-truth base velocity and contact states, respectively, $\mathrm{BCE}$ denotes the binary cross-entropy loss, and $g_{\text{fwd}}$ is a training-only decoder that predicts the next observation $\mathbf{s}_{t+1}$.
We further define the body-state feature $\mathbf{b}_t = [\mathbf{s}_t, \mathrm{sg}[\hat{\mathbf{v}}_t]]$, which augments the observation with the estimated base velocity, where $\mathrm{sg}[\cdot]$ denotes stop-gradient.
This feature conditions both the SMM and ISM on the robot's current motion.

\paragraph{SMM.}
Walking animals retain information about obstacles that have left their field of view to guide subsequent movements~\cite{mcvea2006long,matthis2018gaze}.
Similarly, our SMM integrates visual observations over time to maintain a memory of the surrounding terrain, even as the robot looks in different directions.
At each step, a CNN-based encoder $f_{\text{mem}}$ maps the depth image $\mathbf{D}_t$ to $N$ spatial tokens $\mathbf{T}^{\text{mem}}_t \in \mathbb{R}^{N \times d}$ with learnable positional embeddings.
These tokens are concatenated with $\mathbf{b}_t$ and processed by self-attention~\cite{vaswani2017attention}; the output at the $\mathbf{b}_t$ position, concatenated with the mean of the $N$ visual outputs, forms the summary $\tilde{\mathbf{m}}_t$, which a GRU integrates into the memory state:
\begin{equation}
  \tilde{\mathbf{m}}_t\! =\! \mathrm{SelfAttn}\bigl([\mathbf{b}_t; \mathbf{T}^{\text{mem}}_t]\bigr), \,
  \mathbf{m}_t\! = \!\mathrm{GRU}\bigl(\tilde{\mathbf{m}}_t,\!\mathbf{m}_{t-1}\bigr),
  \label{eq:memory_update}
\end{equation}
where $[\cdot;\cdot]$ denotes token-wise concatenation.

To encourage the memory to retain locomotion-relevant terrain, we project $\mathbf{m}_t$ into two latent representations: $\mathbf{z}^{\text{loc}}_t$ for the surrounding terrain and $\mathbf{z}^{\text{foot}}_t$ for the terrain beneath the feet, which often leaves the field of view before the robot steps onto it.
They are supervised through prototype-based alignment $\mathcal{L}_{\text{proto}}$~\cite{caron2020unsupervised} and height regression, respectively:
\begin{equation}
\begin{aligned}
  \mathcal{L}_{\text{mem}} ={}&
  \mathcal{L}_{\text{proto}}\bigl(g_{\text{loc}}(\mathbf{z}^{\text{loc}}_t), e(\mathbf{H}^{\text{loc}}_t)\bigr) \\
  &+ \bigl\|g_{\text{foot}}(\mathbf{z}^{\text{foot}}_t) - \mathbf{H}^{\text{foot}}_t\bigr\|_1,
\end{aligned}
\label{eq:spatial_memory}
\end{equation}
where $\mathbf{H}^{\text{loc}}_t$ and $\mathbf{H}^{\text{foot}}_t$ denote privileged terrain heights sampled around the robot and beneath its feet, respectively.
The encoder $e$ maps $\mathbf{H}^{\text{loc}}_t$ into a corresponding target embedding, while $g_{\text{loc}}$ and $g_{\text{foot}}$ serve as the alignment head and height decoder.

The encoder $e$ and both auxiliary heads are used only during training.

\vspace{-0.3cm}
\paragraph{ISM.}
In biological sensorimotor systems, nearby space is encoded relative to the body's current configuration and potential actions~\cite{rizzolatti1997space,holmes2004body,marigold2017posterior}.
Inspired by this principle, we introduce the ISM to couple instantaneous visual perception with the robot's physical state and goal command.
Similar to the SMM, an independent encoder $f_{\text{sur}}$ maps the depth image $\mathbf{D}_t$ to $N_{\text{sur}}$ spatial tokens $\mathbf{T}^{\text{sur}}_t = f_{\text{sur}}(\mathbf{D}_t) \in \mathbb{R}^{N_{\text{sur}} \times d}$ with learnable positional embeddings, so that each token corresponds to a region of the visual field.
A cross-attention then uses the body-state feature $\mathbf{b}_t$ as the query and the visual tokens as keys and values:
\begin{equation}
\mathbf{p}_t = \boldsymbol{\alpha}_t W_v \mathbf{T}^{\text{sur}}_t, \; \boldsymbol{\alpha}_t = \mathrm{softmax}\big(\tfrac{(W_q\mathbf{b}_t)(W_k\mathbf{T}^{\text{sur}}_t)^\top}{\sqrt{d}}\big),
\end{equation}
where $W_q$, $W_k$, and $W_v$ are learnable projections and $d$ is the key dimension.
Because $\mathbf{b}_t$ contains proprioception, the estimated base velocity, and the goal command, $\boldsymbol{\alpha}_t$ shifts with the robot's motion and goal direction, so the same observation can yield different representations.

The ISM complements the SMM: while the SMM accumulates terrain information through recurrent integration, the ISM interprets the currently visible scene conditioned on the robot's state and goal, responding directly to changes in visual input.
Unlike the BSM and SMM, which are trained with auxiliary representation-learning objectives, the ISM receives policy gradients directly from the RL objective.
Its encoder and attention are therefore optimized for their contribution to task performance, coupling the extraction of task-relevant visual information with the joint learning of active viewing and whole-body control.
\vspace{-0.3cm}
\paragraph{Actor.}
The actor integrates the three representations with the proprioceptive history to jointly determine body movement and viewing direction.
Its input concatenates $\mathbf{s}_{t-10:t}$ with $\hat{\mathbf{v}}_t$, $\hat{\mathbf{c}}_t$, and $\mathbf{z}^{\text{fwd}}_t$ from the BSM, $\mathbf{z}^{\text{loc}}_t$ and $\mathbf{z}^{\text{foot}}_t$ from the SMM, and $\mathbf{p}_t$ from the ISM.
We feed the SMM latents rather than the raw memory state $\mathbf{m}_t$, so that the actor receives terrain information explicitly grounded by the auxiliary objectives.
An MLP maps this input to the mean of a Gaussian policy:
\begin{equation}
  \mathbf{a}_t \sim \pi_\theta\bigl(\cdot \mid \mathbf{s}_{t-10:t},\, \mathbf{p}_t,\, \mathrm{sg}[\hat{\mathbf{v}}_t, \hat{\mathbf{c}}_t, \mathbf{z}^{\text{fwd}}_t, \mathbf{z}^{\text{loc}}_t, \mathbf{z}^{\text{foot}}_t]\bigr).
  \label{eq:actor}
\end{equation}
The stop-gradient keeps the BSM and SMM shaped solely by their auxiliary objectives, stabilizing their estimates, while the ISM is optimized directly by the policy gradient.

\subsection{Training}
\label{sec:method_training}

\begin{table*}[t]
    \centering
\caption{Simulation performance and ablations (success rate in \%; higher is better).
Values are mean $\pm$ sample standard deviation over three evaluation task sets shared across all methods.
\textbf{Bold} and \underline{underlined} values denote the best and second-best means, respectively.
``Overall'' is the mean over all seen terrains; L9 is the highest difficulty level.}
    \label{tab:ablation_selected_three_rounds}
    \resizebox{\textwidth}{!}{
        \begin{tabular}{@{}l*{10}{c}@{}}
        \toprule
        Method
        & Overall & L9 & Unseen
        & Rough & Boxes & Flat
        & Stairs up & Stairs down & Slope up & Slope down \\
        \midrule

        Hierarchical
       & \secondmeanstd{88.21}{0.47} & \secondmeanstd{85.43}{1.31} & \meanstd{74.13}{0.67}
       & \meanstd{87.73}{1.10} & \meanstd{84.80}{0.87} & \meanstd{95.20}{0.40}
       & \secondmeanstd{82.20}{0.20} & \meanstd{86.73}{0.31}
       & \meanstd{87.80}{0.60} & \meanstd{93.00}{2.12} \\
        
        LoGo-Fh
        & \meanstd{88.02}{0.21} & \meanstd{77.43}{0.29} & \secondmeanstd{78.38}{0.80}
        & \meanstd{85.93}{1.36} & \meanstd{84.93}{2.00} & \meanstd{85.13}{2.14}
        & \meanstd{63.00}{0.00} & \secondmeanstd{99.00}{0.20}
        & \secondmeanstd{99.00}{0.35} & \meanstd{99.13}{0.42} \\

        LoGo-Fh (WFOV-S)
        & \meanstd{19.63}{0.20} & \meanstd{8.86}{0.86} & \meanstd{4.82}{0.52}
        & \meanstd{35.13}{1.81} & \meanstd{13.60}{2.12} & \meanstd{6.93}{0.50}
        & \meanstd{0.33}{0.31} & \meanstd{8.60}{0.60}
        & \meanstd{0.13}{0.12} & \meanstd{72.67}{1.67} \\

        LoGo-Fh (WFOV-D)
        & \meanstd{63.14}{0.51} & \meanstd{39.14}{0.86} & \meanstd{44.51}{1.64}
        & \meanstd{49.73}{0.42} & \meanstd{30.20}{1.00} & \meanstd{38.93}{1.03}
        & \meanstd{67.93}{0.99} & \meanstd{71.53}{1.03}
        & \meanstd{92.73}{2.48} & \meanstd{90.93}{1.30} \\

        LoGo-ISM-MLP
        & \meanstd{71.24}{1.03} & \meanstd{58.57}{3.43} & \meanstd{58.89}{0.80}
        & \meanstd{78.67}{1.47} & \meanstd{76.67}{3.10} & \meanstd{62.53}{0.12}
        & \meanstd{53.80}{2.16} & \meanstd{72.53}{2.23}
        & \meanstd{75.53}{4.35} & \meanstd{78.93}{0.90} \\

        LoGo-SMM-MLP
        & \meanstd{66.23}{0.97} & \meanstd{53.24}{3.15} & \meanstd{48.33}{1.19}
        & \meanstd{54.20}{1.91} & \meanstd{37.27}{3.10} & \meanstd{31.40}{0.87}
        & \meanstd{57.33}{0.42} & \meanstd{96.33}{0.12}
        & \meanstd{91.53}{2.14} & \meanstd{95.53}{1.03} \\

        LoGo w/o AMP
        & \meanstd{85.20}{0.25} & \meanstd{64.10}{1.41} & \meanstd{68.16}{0.58}
        & \bestmeanstd{99.53}{0.12} & \secondmeanstd{95.07}{1.21} & \bestmeanstd{99.87}{0.23}
        & \meanstd{46.60}{0.72} & \meanstd{63.33}{0.58}
        & \meanstd{92.00}{0.35} & \bestmeanstd{100.00}{0.00} \\

        LoGo-SC
        & \meanstd{0.00}{0.00} & \meanstd{0.00}{0.00} & \meanstd{0.00}{0.00}
        & \meanstd{0.00}{0.00} & \meanstd{0.00}{0.00} & \meanstd{0.00}{0.00}
        & \meanstd{0.00}{0.00} & \meanstd{0.00}{0.00}
        & \meanstd{0.00}{0.00} & \meanstd{0.00}{0.00} \\

        \midrule

        LoGo
        & \bestmeanstd{98.47}{0.13} & \bestmeanstd{97.33}{0.16} & \bestmeanstd{93.40}{0.31}
        & \secondmeanstd{97.67}{0.64} & \bestmeanstd{96.13}{0.61} & \secondmeanstd{96.27}{1.10}
        & \bestmeanstd{99.73}{0.23} & \bestmeanstd{99.80}{0.00}
        & \bestmeanstd{99.87}{0.12} & \secondmeanstd{99.80}{0.20} \\

        \bottomrule
    \end{tabular}
    }
    \vspace{-0.5cm}
\end{table*}

We optimize the policy using PPO~\cite{schulman2017proximal}, with rewards divided into four different groups: task, safety, regulation, and style. 
Task rewards encourage the robot to move toward its goal, while safety rewards penalize collisions and falls. 
Regulation rewards promote stable and efficient locomotion, while style rewards encourage natural movements through an adversarial motion prior (AMP)~\cite{peng2021amp} trained on walking and stair-climbing reference motions using terrain-dependent discriminators.

To accommodate the different scales and densities of these reward groups, we adopt a multi-critic architecture~\cite{mysore2022multi} within an asymmetric actor--critic framework~\cite{pinto2017asymmetric}.
Each reward group $k$ has an independent critic $V_k$ that receives privileged state information during training.
Group-specific advantages $\hat{A}^{k}_t$ are computed from the corresponding rewards and value estimates using generalized advantage estimation (GAE)~\cite{schulman2015high}.
These advantages are then normalized and combined to update the shared policy:
$A_t = \sum_{k} w_k \, \mathrm{norm}\left(\hat{A}^{k}_t\right),
$
where $w_k$ controls the contribution of each reward group. 

Importantly, no explicit reward or supervision specifies where the robot should look.
Head movements are subject only to standard action regularization, and the head joint is excluded from the motion discriminator, allowing active viewing to emerge from task-aware policy learning.

\begin{figure}[t]
    \centering
    \includegraphics[width=\columnwidth]{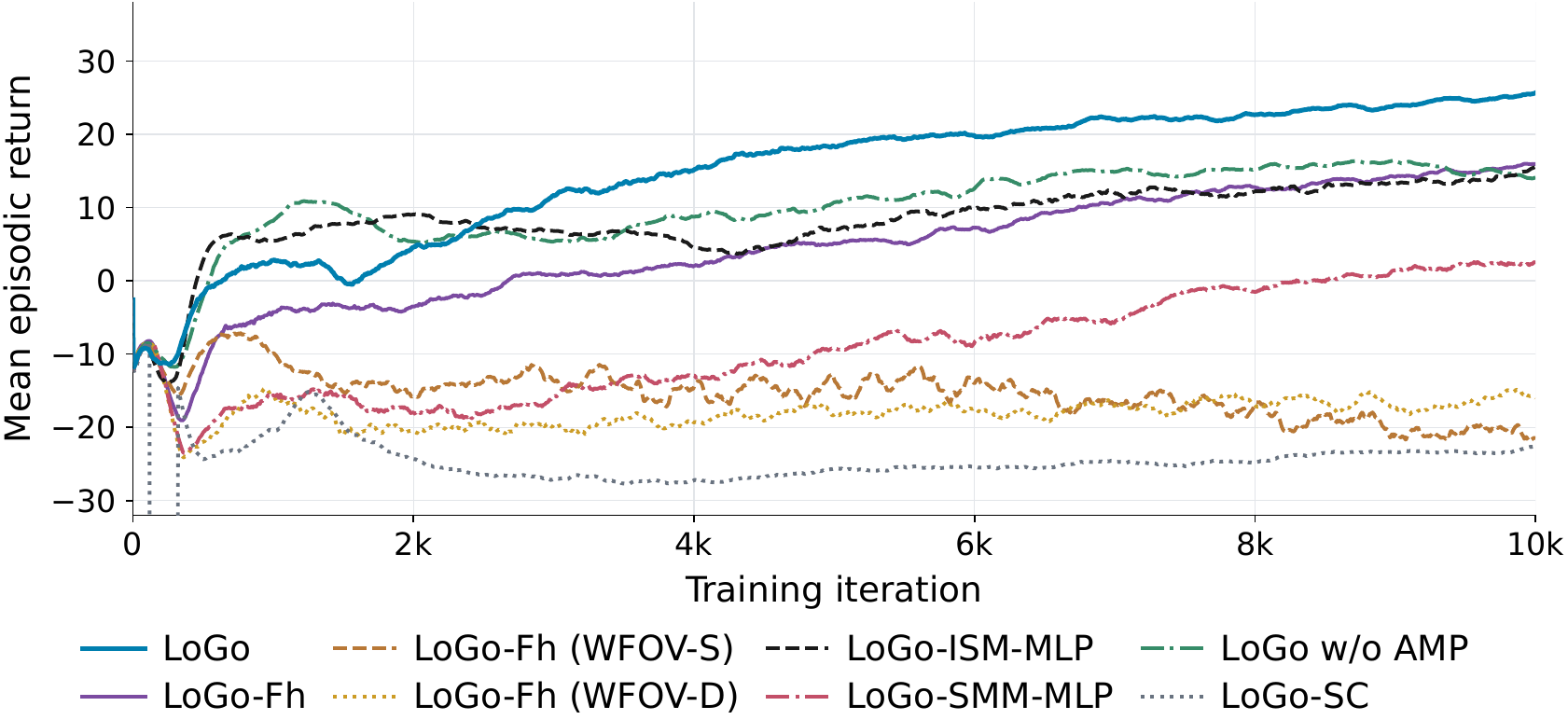}
    \caption{Training performance of LoGo and its ablation variants over 10,000 iterations.
    Curves show the mean episodic return during the first training stage.
    Higher episodic returns generally indicate better learning performance.
    }
    \label{fig:training_curves}
    \vspace{-0.5cm}
\end{figure}

\subsection{Training Setup and Curriculum}
\label{sec:method_env}
\paragraph{Training Setup.}
We train the policy in procedurally generated environments with diverse terrain conditions, including flat, rough, and box-covered ground, stairs, and slopes. 
Goals are preferentially sampled outside the robot's initial field of view, encouraging the policy to actively explore its surroundings and find feasible routes. 
For stairs and slopes, goals are placed on the opposite side of the terrain, encouraging the robot to traverse these challenging surfaces.

During training, we gradually increase the initial heading offset relative to the goal direction and adjust terrain difficulty based on the robot's recent success rate. 

\vspace{-0.5cm}
\paragraph{Progressive training.}
To facilitate stable learning and robust deployment, we train the policy in three stages, each building on the previous one. 
In Stage 1, the robot learns to navigate across different terrains using an ideal depth sensor. 
Stage 2 strengthens motion regularization to encourage smoother and more natural locomotion. 
Stage 3 introduces sensor imperfections to improve real-world robustness, including randomized image capture timing, transmission delays, depth noise, pixel dropout, and small angular offsets.

\section{Experiments and Results}
\label{sec:exp}
We conduct simulation and real-world experiments to address three questions:
\begin{enumerate}[label=\textbf{Q\arabic*:}, leftmargin=*]
    \item In what ways does active perception help goal-directed mobility compared with passive visual sensing?
    \item How do the proposed architectural components and training strategies support learning effective task-aware active perception?
    \item What task-aware viewing behaviors emerge, and how do they relate to subsequent navigation decisions?
\end{enumerate}
\subsection{Simulation Experiments}
\paragraph{Experimental setup.}
\begin{figure*}[t]
    \centering
    \includegraphics[width=\textwidth]{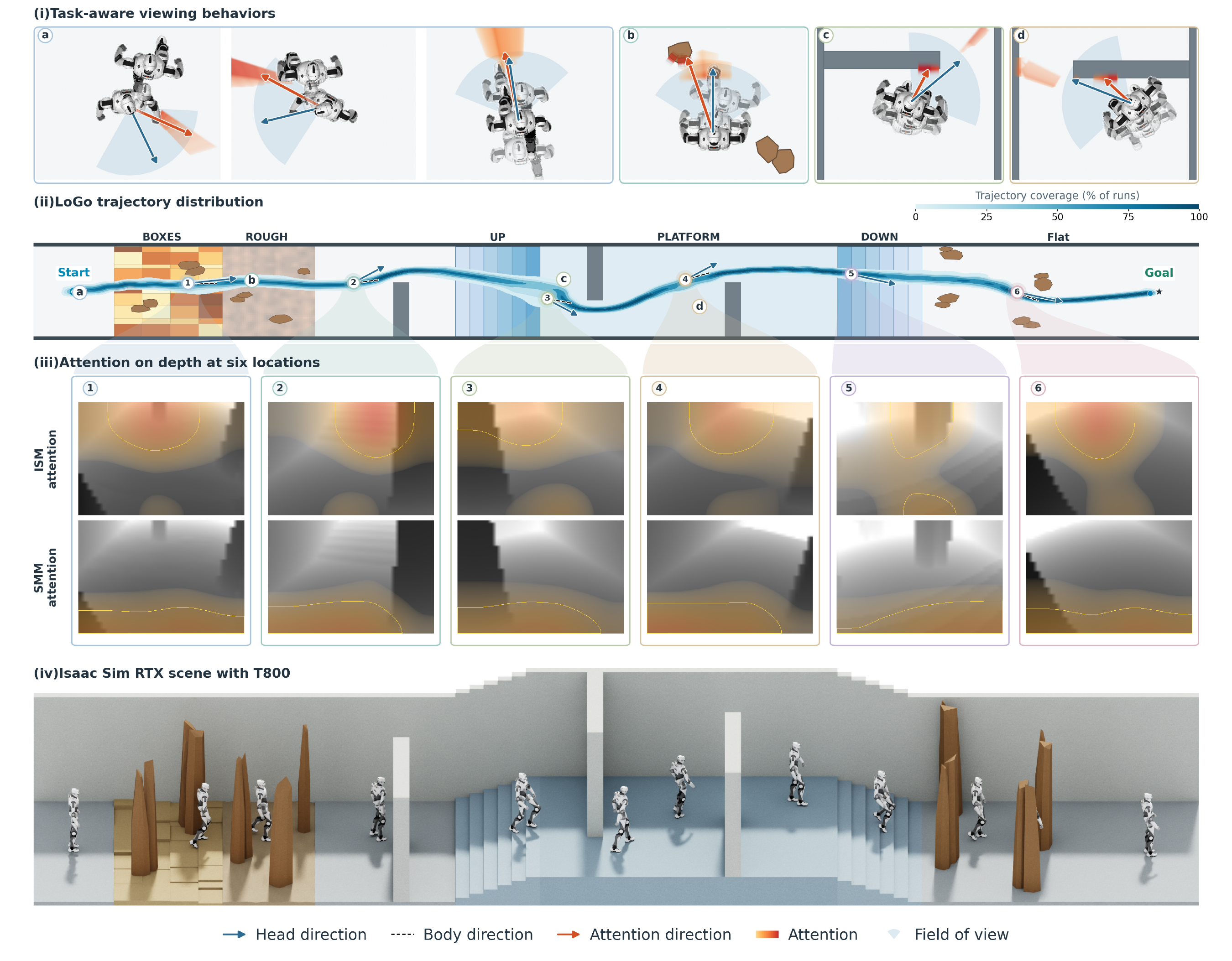}
\vspace{-0.8cm}
\caption{
\textbf{Emergent task-aware viewing during long-horizon navigation.}
(i) Head orientation (blue) and ISM attention direction (orange) in relatively open areas (a) and near obstacles (b--d).
(ii) Trajectory coverage across 100 successful trials, with darker blue indicating higher coverage. 
(iii) Attention overlaid on depth observations: ISM attention (top row) emphasizes structures ahead, while SMM attention (bottom row) focuses on nearby ground. 
(iv) The T800 humanoid traverses a 29\,m evaluation course with varied terrain and obstacles.
}
    \label{fig:longrange_attention}
    \vspace{-0.5cm}
\end{figure*}
We compare \textbf{LoGo} against a hierarchical baseline, three passive-sensing variants, and four ablation variants to examine the benefits of end-to-end joint learning and active perception (Q1), as well as the roles of the attention mechanisms in the representation modules and of key training strategies (Q2).
The \textbf{Hierarchical} baseline~\cite{yang2026spatially} pairs a depth-based navigation policy with a pretrained locomotion policy. The navigation policy outputs velocity commands, which the locomotion policy tracks by producing joint-level actions.
To assess active perception, \textbf{LoGo-Fh} fixes the head at its nominal orientation and removes head actions. \textbf{LoGo-Fh (WFOV-S)} and \textbf{LoGo-Fh (WFOV-D)} further expand the field of view from $70^\circ \times 48^\circ$ to $210^\circ \times 108^\circ$ (horizontal $\times$ vertical), preserving the original image dimensions and angular resolution, respectively.
To assess the role of attention in the representation modules, \textbf{LoGo-ISM-MLP} and \textbf{LoGo-SMM-MLP} replace the cross-attention in the ISM and the self-attention in the SMM with MLPs, respectively, while maintaining the same output dimensions.
For training strategies, \textbf{LoGo w/o AMP} removes the AMP style rewards, while \textbf{LoGo-SC} replaces the multi-critic architecture with a single critic estimating the total reward.
All methods are trained for 10,000 iterations using 16,384 parallel environments in Stage 1.
We evaluate them on seven seen terrain types and three unseen scenarios involving mixed obstacle fields and obstacle-laden stairs.
During evaluation, actions are deterministic and observations are noise-free.
An episode is successful if, within the 25\,s time limit, the robot reaches the goal and stays within 0.5\,m of it for 3\,s, without falling or making invalid body contacts.
\vspace{-0.5cm}
\paragraph{Effectiveness of Active Perception (Q1).}
LoGo outperforms the \textbf{Hierarchical} baseline in overall success (98.47\% vs.\ 88.21\%), with larger margins at the highest difficulty level L9 (97.33\% vs.\ 85.43\%) and in unseen scenarios (93.40\% vs.\ 74.13\%), suggesting the benefit of joint learning. To isolate the role of active perception, we further compare LoGo with its passive-sensing variants. LoGo outperforms all of them on every seen terrain type and in unseen scenarios. Compared with LoGo-Fh, the gains are largest in stair ascent (99.73\% vs.\ 63.00\%) and at L9 (97.33\% vs.\ 77.43\%), and extend to unseen terrain combinations (93.40\% vs.\ 78.38\%).
Expanding the field of view does not close this gap: both wide-field-of-view variants perform substantially worse than LoGo-Fh, and their training returns stagnate at low values (Fig.~\ref{fig:training_curves}), indicating that reduced angular resolution or enlarged visual input makes learning considerably harder.
These results suggest that active viewing is more effective than passively expanding visual coverage.

\paragraph{Contributions of Architectural and Training Designs (Q2).}
Prior work has shown the importance of estimating both the robot's state and surrounding terrain for challenging locomotion~\cite{luo2024pie}.
Building on this insight, we further investigate the effects of attention mechanisms and learning strategies in LoGo.
Replacing cross-attention in the ISM or self-attention in the SMM with MLPs reduces overall success from 98.47\% to 71.24\% and 66.23\%, respectively (Table~\ref{tab:ablation_selected_three_rounds}). The ISM result supports the importance of cross-attention for aggregating current visual information conditioned on the body state. For the SMM, replacing self-attention while preserving recurrent integration highlights the contribution of attention-based visual processing within the memory pathway. Both variants exhibit larger performance drops in unseen scenarios and slower learning (Fig.~\ref{fig:training_curves}), suggesting that these attention mechanisms contribute to generalization and training efficiency. 

Removing AMP primarily impairs stair traversal (46.60\% vs. 99.73\% on stair ascent). 
Replacing the multi-critic design with a single critic causes training to fail (0\% success in all scenarios) and yields the lowest return throughout training (Fig.~\ref{fig:training_curves}).
We believe this is because the dense regulation and style rewards dominate the single critic's value estimate, making it harder to learn from the sparse task reward. 
Beyond these experiments, we evaluate LoGo on the JS01 quadruped robot, achieving an overall success rate of 98.26\% across the same seven terrains.

\paragraph{Emergent Task-Aware Viewing Behaviors (Q3).}
\begin{figure}[t]
    \centering
    \includegraphics[width=\columnwidth]{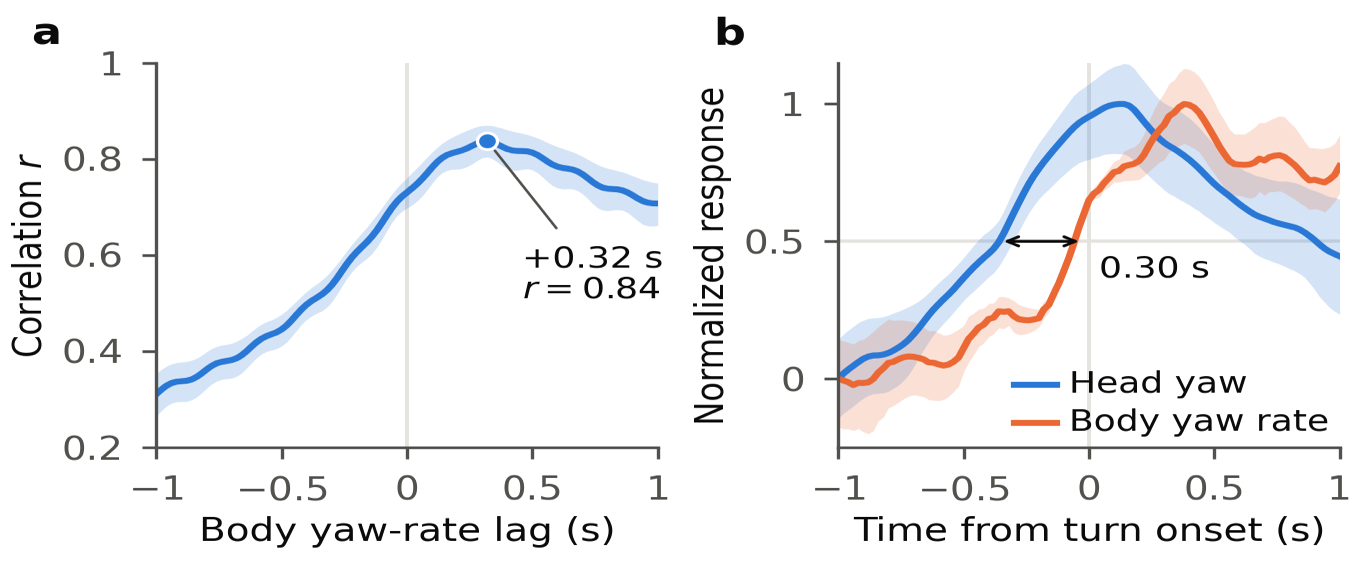}
    \vspace{-0.5cm}
    \caption{
    \textbf{Head reorientation precedes body turning.}
    (a) Lagged correlation between head yaw and body yaw rate.
    (b) Mean normalized responses across 21 turning events, aligned to body-turn onset; the arrow indicates the half-amplitude lead.
    Shading indicates 95\% confidence intervals from obstacle-field bootstrapping.
    }
    \label{fig:head_body_evidence}
\end{figure}

To investigate the viewing behaviors learned by LoGo, we analyze how viewing direction and visual attention adapt to changing task demands during goal-directed mobility.
As illustrated in Fig.~\ref{fig:longrange_attention}(i), when the surroundings are relatively unobstructed, the robot tends to orient its head toward the goal, while the ISM attention emphasizes goal-aligned visual regions (a).
In such conditions, viewing direction and visual attention are broadly aligned, supporting goal-directed progression.
In contrast, when nearby obstacles constrain the available path, the two exhibit complementary behaviors: the robot redirects its head toward potential routes and openings, while the ISM attention emphasizes visible obstacle boundaries and geometric constraints relevant to safe passage (b--d).
This suggests that active viewing and visual attention play complementary roles, with the former acquiring information about possible routes and the latter identifying constraints within the observed scene.
Their coordination shifts with the immediate demands of the task rather than following a fixed relationship, illustrating how task-aware active perception manifests in behavior.

To further evaluate these emergent viewing behaviors during sustained mobility, we design a challenging 29\,m long-horizon course integrating rough and box terrains, ascending and descending stairs, staggered walls, and obstacle fields (Fig.~\ref{fig:longrange_attention}(ii) and (iv)).

Across 100 trials with initial position perturbations of up to 0.15\,m and an extended time limit of 60\,s, LoGo completes the course in every trial, with completion times ranging from 38 to 46\,s.
The trajectories remain closely clustered, while the robot dynamically adjusts its head orientation along the route.
At six representative locations, the ISM primarily attends to structures ahead, while the SMM focuses on nearby ground (Fig.~\ref{fig:longrange_attention}(iii)), revealing complementary attention patterns during continuous mobility.

\paragraph{Relationship Between Viewing and Subsequent Actions (Q3).}
\begin{figure}[t]
    \centering
    \includegraphics[width=\columnwidth]{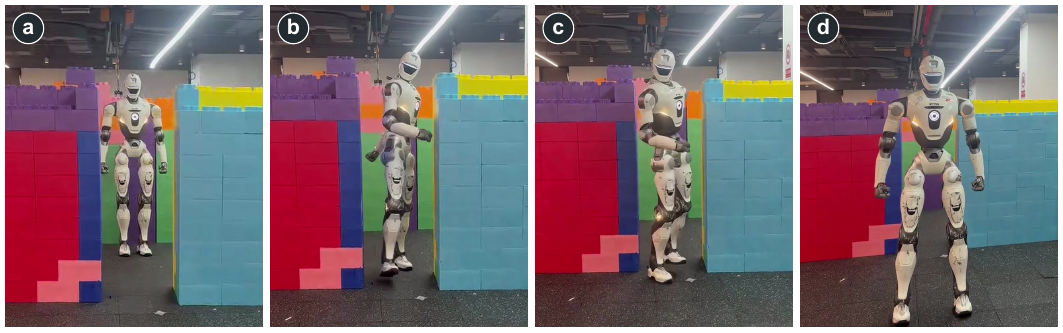}
\caption{\textbf{Coordinated viewing and body orientation in a narrow passage.}
LoGo enables the T800 humanoid to traverse a 60\,cm-wide passage by orienting its body sideways while keeping its viewing direction along the passage, maintaining a view of the route ahead throughout sideways movement.}
\label{fig:narrow_passage}
\end{figure}
\begin{figure*}[t!]
    \centering
    \includegraphics[width=\linewidth]{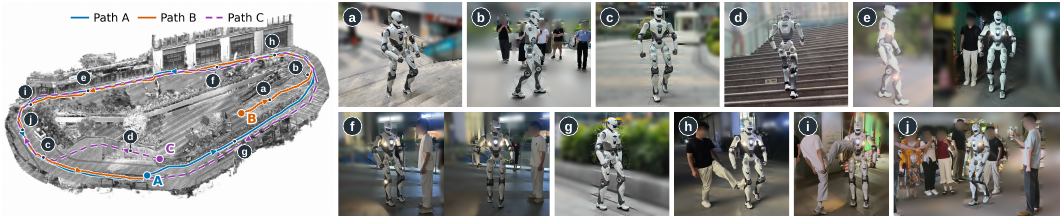}
    \caption{
        \textbf{Real-world evaluation of LoGo on the T800 humanoid.}
        The map shows three outdoor evaluation routes (Paths A--C), with labeled locations corresponding to the example scenes.
        The robot ascends stairs (a, d), walks along paved paths (c, g), navigates around pedestrians (b, f, j), and operates under different lighting conditions (e).
        It also experiences manually applied perturbations to the leg (h) and torso (i), after which it regains balance and continues walking.
    }
    \label{fig:experiment}
\end{figure*}
Beyond determining where to look, we investigate how active viewing relates to subsequent body movements.
As shown in Fig.~\ref{fig:head_body_evidence}, LoGo exhibits an anticipatory viewing pattern in which head reorientation precedes body turning.
The lagged correlation between head yaw relative to the body and body yaw rate peaks when head orientation leads by 0.32\,s ($r=0.84$).
Consistent with this temporal relationship, across 21 turning events, the head has already rotated an average of $31^\circ$ toward the upcoming turning direction when the body turning begins.

To examine whether this relationship merely reflects their shared dependence on the navigation goal, we further analyze future body heading changes while accounting for goal bearing and current body rotation.
Including head orientation improves held-out prediction accuracy, increasing $R^2$ from 0.78 to 0.84.
These results suggest that head orientation carries information about upcoming body movements beyond the current task command.
Together, these findings exemplify the emergent Look-to-Go behavior, in which the robot proactively redirects its viewing direction ahead of subsequent body movements, highlighting the learned temporal coordination between active perception and whole-body locomotion.

\subsection{Real-World Experiments}

We deploy LoGo on the full-size T800 humanoid robot and evaluate its performance in both indoor and outdoor environments.
Our experiments examine three aspects of real-world mobility: navigating confined spaces, maintaining reliable performance across repeated outdoor trials under changing environmental conditions, and responding to visual occlusions and external disturbances.

\noindent\textbf{Indoor Evaluation.} In the laboratory, we construct narrow passages and occluded corners using obstacle walls to evaluate mobility under restricted visibility and limited clearance.
In the serpentine layout shown in the upper part of Fig.~\ref{fig:intro}, the robot actively adjusts its viewing direction to inspect upcoming sections of the route as it progresses through successive turns.
In the 60\,cm-wide passage shown in Fig.~\ref{fig:narrow_passage}, it turns its body sideways to accommodate the limited clearance while keeping its view directed along the passage.
This allows the robot to observe the route ahead even when its body orientation differs from its direction of travel.
Together, these behaviors demonstrate flexible coordination between active viewing and whole-body movement in response to route geometry and spatial constraints.

\noindent\textbf{Outdoor Evaluation.} We establish three distinct inspection routes (Paths A--C) within the same urban environment, as illustrated in Fig.~\ref{fig:experiment}.
Each trial covers approximately 170--200\,m, with 10 trials conducted per route at different times, including daytime and nighttime, resulting in 30 trials in total.
The three routes expose the robot to different sequences of terrain features and spatial constraints, including stairs and pedestrian-populated areas.
Although the underlying environment remains the same, the robot encounters naturally varying illumination, pedestrian traffic, and transient changes in the surrounding scene.
In crowded sections, dense pedestrian traffic creates frequent visual occlusions and continually changes the space available for movement, posing substantial challenges for reliable mobility.
A representative behavior occurs when a pedestrian occludes the robot's view of the route ahead.
The robot first actively turns its head to inspect the surrounding space for a feasible passage.
After advancing along the route, it redirects its view toward the pedestrian to reassess their relative position and maintain clearance while moving.
This sequence shows how LoGo adjusts its viewing direction to changing task demands, shifting from finding a passage to tracking a nearby moving obstacle.
Despite these challenges, LoGo completes all 30 trials ($10/10$ per route), achieving a 100\% success rate over approximately 5.1--6.0\,km of cumulative outdoor traversal.
These results show that LoGo maintains reliable mobility across different outdoor routes and changing environmental conditions.

We further evaluate robustness to visual occlusions and external physical disturbances.
When the robot's visual observations are deliberately obstructed during movement, it maintains stable locomotion and actively redirects its head toward less occluded regions.
Under externally applied pushes and kicks, the robot recovers its balance and continues moving, as illustrated in Fig.~\ref{fig:experiment}.
Together, these experiments demonstrate the practical capability of LoGo to coordinate active perception and body movements in dynamic, human-centered environments.

\section{Conclusion}
\label{sec:conclu}
We presented LoGo, an end-to-end RL framework that jointly learns task-aware active perception, goal-directed navigation, and whole-body locomotion for legged mobility. 
By integrating Body State, Spatial Memory, and Immediate Surrounding representations, LoGo learns to acquire task-relevant visual information in anticipation of subsequent actions, without explicit viewpoint supervision or dedicated observation rewards.
Our experimental results demonstrate the value of jointly learning visual information acquisition and physical control. 
These findings suggest that purposeful active perception can emerge from learning to act, offering a path toward more adaptive and autonomous legged mobility. 

However, our evaluation does not include discontinuous terrain or environments with low overhead clearance.
Future work will explore how to better coordinate active perception and whole-body control for precise foothold selection and adaptive adjustments of body height and posture, extending LoGo's mobility to these challenging settings.

\section*{LLM Usage Statement}

We used ChatGPT for language polishing and preliminary figure design, and Claude for coding. All AI-assisted content was reviewed and verified by the authors.

{
    \small
    \bibliographystyle{ieeenat_fullname}
    \bibliography{main}
}
\clearpage
\appendix
\raggedbottom

\setcounter{topnumber}{3}
\setcounter{bottomnumber}{2}
\setcounter{totalnumber}{5}
\setcounter{dbltopnumber}{2}
\renewcommand{\topfraction}{0.95}
\renewcommand{\bottomfraction}{0.85}
\renewcommand{\textfraction}{0.05}
\renewcommand{\floatpagefraction}{0.75}
\renewcommand{\dbltopfraction}{0.95}
\renewcommand{\dblfloatpagefraction}{0.75}
\setlength{\textfloatsep}{10pt plus 2pt minus 2pt}
\setlength{\floatsep}{8pt plus 2pt minus 2pt}
\setlength{\intextsep}{10pt plus 2pt minus 2pt}
\setlength{\dbltextfloatsep}{10pt plus 2pt minus 2pt}
\setlength{\dblfloatsep}{8pt plus 2pt minus 2pt}
\setlength{\emergencystretch}{1em}
\clubpenalty=10000
\widowpenalty=10000
\displaywidowpenalty=10000
\setlength{\abovecaptionskip}{5pt}
\setlength{\belowcaptionskip}{0pt}
\ifdefined\captionsetup
  \captionsetup[table]{position=top,skip=4pt}
  \captionsetup[figure]{position=bottom,skip=5pt}
\fi

\end{document}